\documentclass[runningheads]{llncs}
\usepackage[T1]{fontenc}
\usepackage{graphicx}
\usepackage{subcaption}
\usepackage{cite}
\usepackage{acro}
\usepackage{float}
\usepackage{xcolor}
\usepackage{amsmath}
\usepackage{multirow}
\usepackage{multicol}
\usepackage{wrapfig}
\usepackage{booktabs}  
\usepackage{subcaption}
\usepackage{wrapfig}
\usepackage{float}
\usepackage[english]{babel}
\usepackage{hyperref}
\usepackage{adjustbox}
\usepackage{amssymb}
\usepackage{tabularray}
\usepackage{color}

\begin{document}
\title{Multi-Modal Mamba Modeling for Survival Prediction (M4Survive): Adapting Joint Foundation Model Representations}
%
%
\author{Ho Hin Lee\inst{1} \and
Alberto Santamaria-Pang\inst{1*} \and
Jameson Merkov\inst{1} \and
Matthew Lungren\inst{1} \and
Ivan Tarapov\inst{1}}
%
\authorrunning{Ho Hin et al.}

\institute{Microsoft Health AI, Redmond WA , USA \\
\email{alberto.santamariapang@microsoft.com}\\
}
\maketitle              
\begin{abstract}
Accurate survival prediction in oncology requires integrating diverse imaging modalities to capture the complex interplay of tumor biology. Traditional single-modality approaches often fail to leverage the complementary insights provided by radiological and pathological assessments. In this work, we introduce M4Survive (Multi-Modal Mamba Modeling for Survival Prediction), a novel framework that learns joint foundation model representations using efficient adapter networks. Our approach dynamically fuses heterogeneous embeddings from a foundation model repository (e.g., MedImageInsight, BiomedCLIP, Prov-GigaPath, UNI2-h), creating a correlated latent space optimized for survival risk estimation. By leveraging Mamba-based adapters, M4Survive enables efficient multi-modal learning while preserving computational efficiency. Experimental evaluations on benchmark datasets demonstrate that our approach outperforms both unimodal and traditional static multi-modal baselines in survival prediction accuracy. This work underscores the potential of foundation model-driven multi-modal fusion in advancing precision oncology and predictive analytics. Both codes and interactive demo are available at: https://github.com/microsoft/healthcareai-examples
\end{abstract}

\section{Introduction}
Survival prediction in oncology requires the integration of multi-scale imaging data to capture the complex interplay of biological and clinical variables. Radiology offers macro-level structural insights, while pathology provides detailed micro-level cellular information. However, traditional unimodal approaches fail to leverage the complementary prognostic value embedded within these disparate data sources, and classical radiomics methods often struggle to correlate features across these modalities, thereby limiting their capacity to model tumor heterogeneity and predict treatment response. \par
Foundational reviews by Lambin et al. and Aerts et al. have demonstrated how quantitative image analysis can inform risk prediction, while also highlighting the inherent limitations of single-modality approaches \cite{Lambin2016, Aerts2020}. Similarly, surveys on deep multi-modal learning and segmentation methods reveal that, although current techniques show promise, they frequently lack efficient cross-scale integration and computational scalability  \cite{Wang2021, Garcia2022}. \par
Recent advances in multi-modal imaging analysis have shown that aggregating information from different modalities can significantly enhance model performance. For instance, transformer-based models have been developed that integrate radiological, pathological, and molecular data for glioblastoma survival prediction, demonstrating substantial improvements over unimodal techniques despite challenges such as increased computational complexity and interpretability issues \cite{gomaa2024comprehensive}. Similar studies in colon cancer have fused pathology and genomic data, further highlighting both the potential and the inherent difficulties of integrating heterogeneous sources, especially when accounting for temporal variations and data heterogeneity \cite{Smith2023}. \par
Despite these promising developments, existing methods predominantly operate under a “complete data” paradigm, wherein every patient record is assumed to have all modalities available \cite{chen2020pathomic, braman2021deep, Lee2023, Zhang2023}. In routine clinical practice, however, missing modalities are common—leading to data scarcity and an elevated risk of overfitting when incomplete cases are discarded. Moreover, static multi-modal fusion strategies have proven insufficient for fully capturing the dynamic and cross-scale interactions between radiology and pathology data \cite{Chen2022}. Other approaches, such as modality dropout and uni-modal feature reconstruction, have attempted to mitigate these issues but often fall short of effectively leveraging the available data and thoroughly comparing model performance across scenarios with and without missing modalities \cite{cui2022survival, liu2022moddrop++}. \par
To address these challenges, we introduce M4Survive (Multi-Modal Mamba Modeling for Survival Prediction), a novel framework that learns joint representations from foundation models trained on radiology and pathology data. By leveraging Mamba-based adapter networks, M4Survive fuses heterogeneous embeddings from multiple pre-trained models (e.g., MedImageInsight, BiomedClip, Prov-GigaPath, UNI2-h) into a correlated latent space optimized for survival risk estimation. The key contributions of our work are:
\begin{itemize}
    \item \textbf{Adaptive Multi-Modal Fusion:} Our framework employs efficient adapter networks to dynamically integrate multi-scale imaging data. This approach accommodates variations in data availability and effectively captures the complex interactions between macro-level and micro-level features—ensuring robustness in real-world scenarios.
    \item \textbf{Enhanced Predictive Accuracy and Interpretability:} By jointly analyzing radiology and pathology data, M4Survive outperforms traditional unimodal and static multi-modal baselines. The resulting correlated latent space not only improves predictive accuracy but also enhances interpretability, offering deeper insights into how different imaging modalities contribute to survival prediction.
    \item \textbf{Scalability and Clinical Applicability:} Although our current focus is on radiology and pathology, the adapter-based architecture of M4Survive is inherently scalable and only needs ~15 seconds for training with small amount of CPU/GPU memory.
\end{itemize}

%
\begin{figure}[t!]
    \centering
     \includegraphics[width=\columnwidth]{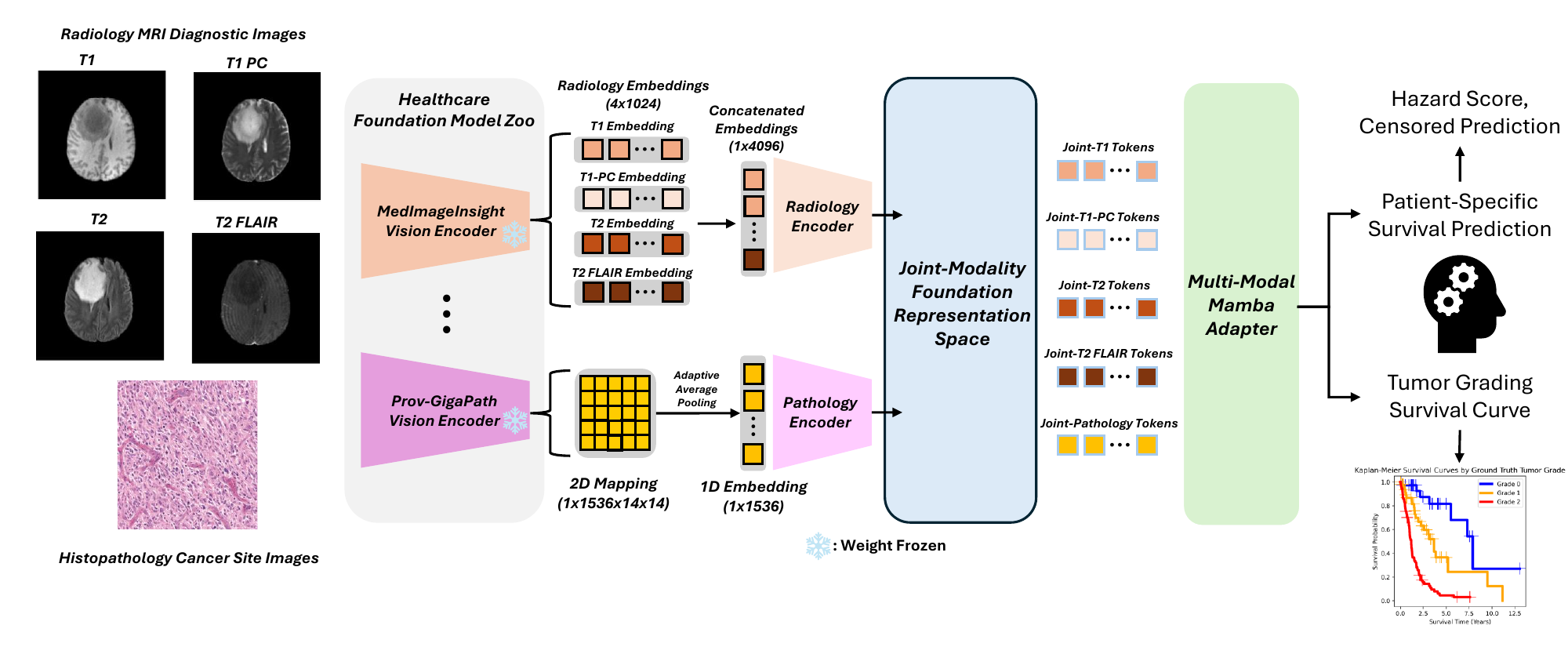}
    \caption{Overview of M4Survive. MRI diagnostic images with variable contrast and cancer site histopathology images are independently processed through dedicated healthcare foundation models to generate modality-specific embeddings. Such embeddings are subsequently mapped into a joint-modality semantic space following with modality-specific encoders and leverage a Mamba adapter to perform multi-modal fusion for survival predictions.}
\end{figure}

\section{Methods}
In contrast to conventional fusion strategies that simply combine features via element-wise averaging or concatenation, our approach learns a joint foundation semantic space using modality-specific embeddings and an efficient adapter network. Our method follows a two-step hierarchical pipeline for multi-modal fusion. First, we pre-compute image embeddings using variable foundation models and then train modality-specific encoder networks to project these embeddings into a unified semantic space. Second, the resulting joint embeddings are fed as multiple tokens into a lightweight Mamba adapter network, which facilitates fine-grained cross-modal interactions for enhanced survival prediction.
\subsection{Learning Joint Foundation Semantic Space}
Our approach begins by leveraging both general and modality-specific foundation models to extract high-dimensional embeddings from multi-modality images. In our setting, we use 2D diagnostic MRI slides with four different contrasts ($I_{T1}, I_{T1PC}, I_{T2}, I_{FLAIR}$) and 2D cancer site pathology images (i.e., H\&E, denoted as $I_{Path}$) images. It is important to note that M4Survive can scale to accommodate longer sequences of images or 3D imaging, provided that the foundation models can generate the necessary embeddings. For instances, we leverage MedImageInsight and Prov-GigaPath as our radiology-pathology foundation models respectively and compute embeddings as follow:
\begin{equation}
\begin{aligned}
    \langle F_{T1}, F_{T1PC}, F_{T2}, F_{FLAIR} \rangle &= MedImageInsight(\langle I_{T1}, I_{T1PC}, I_{T2}, I_{FLAIR} \rangle) \\
    F_{Path} &= ProvGigaPath(I_{Path})
\end{aligned}
\end{equation}
Since these embeddings originate from distinct feature spaces, we employ small, efficient encoder networks (i.e. 2 layers of Multi-Layer Peceptron (MLP)) to project each modality's features into a unified semantic space as tokens:
\begin{equation}
\begin{aligned}
    T_{Rad} & = MLP_{Rad}(\langle F_{T1}, F_{T1PC}, F_{T2}, F_{FLAIR} \rangle) \\
    T_{Path} & = MLP_{Path}(I_{Path}) 
\end{aligned}
\end{equation}
\noindent Where $T_{rad}\in \mathcal{R}^{4\times Channel_{rad}}$ and $T_{Path}\in \mathcal{R}^{1\times Channel_{Path}}$ are the jointly normalized tokens of all MRI contrast modalities and the pairwise pathology image respectively. This projection step is crucial as it aligns the disparate representations, thereby bridging the semantic gap between the two modalities. By normalizing and compressing the modality-specific embeddings into the same hidden dimension and assume it is the learnable joint semantic space, our method facilitates a coherent and integrative representation that enables the downstream model to effectively reason over complementary information from both imaging domains.

\subsection{Multi-Modal Fusion with Mamba Adapter}
The modality-specific embedding encoders encode the foundation model outputs and concatenate it as a sequence of input tokens for a lightweight Mamba adapter network \cite{gu2023mamba}. Rather than collapsing these features into a single vector via concatenation or averaging, our token-based fusion strategy leverages a selective state-space model to preserve fine-grained, modality-specific details. For each token $t_n\in\mathcal{R}^{d_u}$ in a long sequence concatenating with $T_{rad}$ and $T_{Path}$, where $n=1,...,5$ is the total number of image modalities, the mamba network updates a hidden state $h_n$ using the recurrence and generate an output representation of $y_n$ as follows: 
\begin{equation}
    h_{n+1} = A_dh_n + B(t_n)t_n
\end{equation}
\begin{equation}
    y_n=C(t_n)h_n
\end{equation}
where $A_d\in\mathcal{R}^{d_h\times d_h}$ is a fixed state transition matrix that propagates the accumulated context; $B(t_n)\in\mathcal{R}^{d_h\times d_n}$ is an input-dependent function that modulates the contribution fo the current token $t_n$; $C(t_n)\in\mathcal{R}^{d_y\times d_h}$ is an input-dependent function that extracts a refined output $y_n\mathcal{R}^{d_{y}}$ from the hidden state $h_n$. Unlike transformer network, within this formulation, each token—whether from radiology or pathology—selectively contributes to the evolving state through the term $B(t_n)t_n$, while $A_dh_n$carries forward the existing context. The output $y_n$is then dynamically generated based on the current token and the historical context. This selective state-space update mechanism mirrors the dynamic gating of self-attention in transformers but avoids the quadratic computational complexity by operating sequentially with linear complexity. Consequently, the Mamba adapter efficiently captures complex cross-modal interactions, ensuring that the inter-modal relationships are preserved, thereby enhancing the overall predictive performance for survival. 

\subsection{Survival Modeling}
For survival prediction, we follow previous works \cite{chen2020pathomic, cui2022survival} and employ the Cox ranking loss function to train the adapter model, generating ranked hazard values within the range of -3 to +3 as follows:
\begin{equation}
    \mathcal{L}_{cox}=-\sum_{i:E_{i}=1}(F_{\theta}(m_{i})-\log\sum_{j:s_{i}\geq s_{j}}\exp^{F_{\theta}(m_{i})})
\end{equation}
where $m_i$ represents the joint multi-modal patient representations, $E_i$ denotes the censor status, and $F_{\theta}$ is the Mamba adapter network that predicts patient-specific hazard values (risk of death). A higher hazard value indicates greater relative risk, but these scores do not directly translate into survival probabilities.

To derive time-dependent survival probabilities, we rely on the Cox Proportional Hazards (Cox) model \cite{cox1972regression}. The Cox model expresses the hazard function as:
\begin{equation}
    h(t | m_i) = h_0(t) \exp(F_{\theta}(m_i))
\end{equation}
where $h_0(t)$ is the baseline hazard function, capturing the underlying mortality risk for a reference patient, and $F_{\theta}(m_i)$ acts as a scaling factor, shifting the baseline hazard based on the patient's multi-modal features (e.g., MRI and pathology embeddings). This formulation allows us to estimate survival probabilities over time, providing clinically meaningful risk assessments beyond relative hazard rankings.

\begin{table*}[t]
    \centering
    \begin{adjustbox}{width=0.8\textwidth}
    \begin{tabular}{c|c|c|c}
    \toprule
        Method & Radiology & Pathology & C-Index \\
        \midrule
        Pathomic Fusion \cite{chen2020pathomic} & $\times$ & $\checkmark$ & 77.13$\pm$1.04 \\
        Deep Orthogonal Fusion \cite{braman2021deep} & $\checkmark$ & $\checkmark$ & 75.19$\pm$2.13\\
        \textit{Can et al}. w/out M.D \cite{cui2022survival} & $\checkmark$ & $\checkmark$ & 76.54$\pm$1.32\\ 
        \midrule
        M4Survive (Our approach) & $\checkmark$ & $\checkmark$ & \textbf{81.27$\pm$0.56}\\ 
        \bottomrule
    \end{tabular}
    \end{adjustbox}
    \caption{The Quantitative Evaluation with Current SOTAs for Benchmarking (M.D: Modality Dropout)}
\end{table*}

\section{Experimental Setup}
Inspired by Can et al. \cite{cui2022survival}, we collect a large-scale dataset consisting exclusively of complete glioma tumor data—comprising 1,698 samples from 962 patients—by integrating data from TCGA, TCIA, and the BraTs dataset. We further select the subjects (i.e. 170) that are available with radiology and pathology pairs and accompanied by survival time and censor status. \par
\textbf{Experimental Pipeline.} We first deploy healthcare foundation models through Azure Model AI Catalog and generate image embeddings. For foundation models, we have pre-selected current state-of-the-art (SOTA) general embedding model such as BiomedCLIP \cite{zhang2023biomedclip} and MedImageInsight \cite{codella2024medimageinsight}, and modality-specific models such as Prov-GigaPath \cite{xu2024whole} and UNI2-h \cite{chen2024towards} for evaluation. After that, both the uni-modal encoder models and the adapter network are trained jointly in supervised setting with survival dates and censor status as ground truth label, with defined data split (training: 75\%, validation: 5\%, testing:20\%) To evaluate the effectiveness of our approach, we leverage the best combination result to compare several previous SOTA approaches and benchmark in the Radiology-Pathology only setting \cite{chen2020pathomic,braman2021deep,cui2022survival}. To further understand the capability of adapter network, we further perform ablation studies to investigate the variability of using different network backbones such as MLP and transformer for multi-modal fusion.\par
\textbf{Evaluation.} Survival prediction is evaluated using the concordance index (c-index), a metric that quantifies the predictive accuracy by measuring the proportion of correctly ranked patient pairs based on the survival outcomes (1: perfect prediction, 0.5: representing random chance). Both the modality-specific encoders and adapter network are trained with a batch size of 16, a learning rate of 0.0003, over 30 epochs. Experiments employing MLP and transformer are executed on a 24-core Intel CPU, while NVIDIA V100 GPU is used to train Mamba network. We directly leverage the predicted scores to rank patients and classify the predicted scores into three strata, using the 33rd and 66th percentiles, thereby mapping continuous hazard scores into tumor grading categories \cite{chen2020pathomic}.

\section{Results}
\subsection{Comparison with State-of-the-Art Methods}
Table 1 compares our proposed M4Survive approach with several state-of-the-art (SOTA) fusion methods. Pathomic Fusion and Deep Orthogonal Fusion primarily focus on combining histopathology and genomic data, while the method by Can et.al. integrates radiology and pathology features. In contrast, M4Survive leverages multiple imaging modalities within a unified architecture. As shown, M4Survive achieves a concordance index (c-index) of $81.27\pm0.56$, outperforming current SOTA approaches by 5.37\%. This improvement demonstrates the advantage of incorporating complementary information from all available modalities and highlights the effectiveness of our novel multi-modal fusion design.

\begin{figure}[t]
    \centering
     \includegraphics[width=\columnwidth]{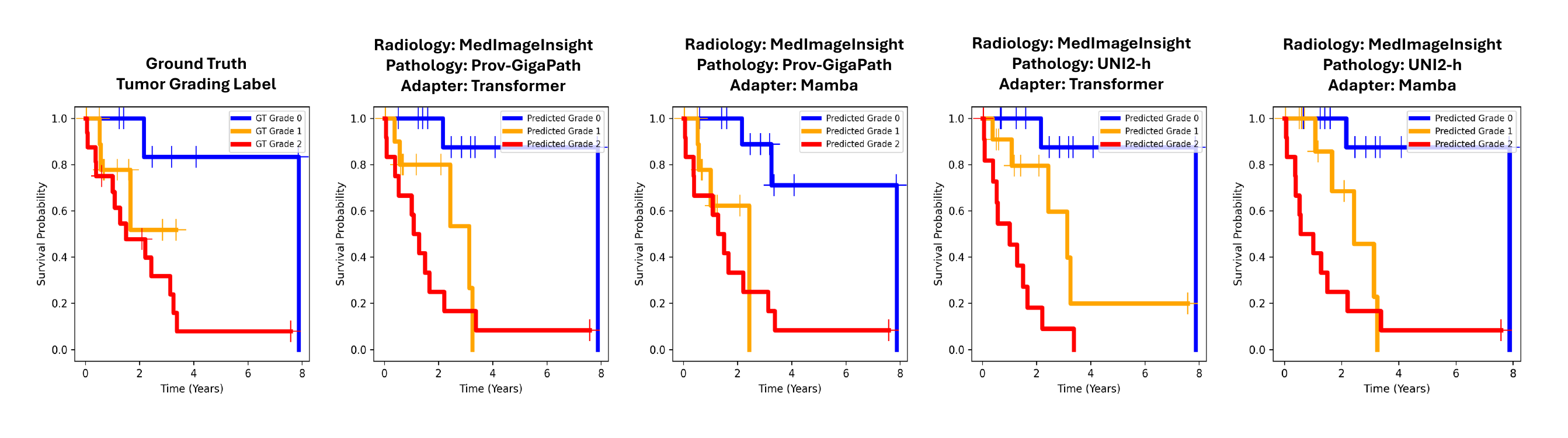}
    \caption{Kaplan–Meier survival curves for high-risk and low-risk groups predicted by two ablation configurations using the Mamba and transformer adapters. }
\end{figure}

\begin{table*}[t]
    \centering
    \begin{adjustbox}{width=0.8\textwidth}
    \begin{tabular}{c|c|c|c|c} 
    \toprule
    \multicolumn{2}{c|}{Foundation Models} & Adapter Network & \multicolumn{2}{|c}{Predictions}  \\
    \midrule
    Radiology & Pathology & Architecture & C-Index & Censored\\
    \midrule
    BiomedCLIP & $\times$ & MLP &  62.31$\pm$5.73 & 59.87$\pm$4.26\\
    MedImageInsight & $\times$ & MLP & 72.51$\pm$3.29& 68.57$\pm$3.79\\
    $\times$ & BiomedCLIP & MLP & 76.59$\pm$3.32 & 72.39$\pm$2.54\\
    $\times$ & Prov-GigaPath & MLP & 77.34$\pm$1.67 & 74.29$\pm$2.43\\
    $\times$ & UNI2-h & MLP & 78.16$\pm$1.45 & 76.17$\pm$3.21\\
    BiomedCLIP & BiomedCLIP & MLP & 74.42$\pm$1.26 & 75.06$\pm$1.08\\
    MedImageInsight & MedImageInsight & MLP & 74.32$\pm$0.42 & 72.38$\pm$2.70\\
    MedImageInsight  & Prov-GigaPath & MLP & 78.50$\pm$0.23 & 80.66$\pm$ 2.57\\
    MedImageInsight & UNI2-h  & MLP & 79.46$\pm$0.10 & 83.67$\pm$ 2.11\\
    \midrule
    BiomedCLIP & BiomedCLIP & Transformer & 75.23$\pm$1.13 & 76.07$\pm$1.23\\
    MedImageInsight  & MedImageInsight & Transformer & 75.43$\pm$0.76 & 76.19$\pm$2.69 \\
    MedImageInsight  & Prov-GigaPath & Transformer & 79.07$\pm$0.87& 78.10$\pm$2.69\\
    MedImageInsight  & UNI2-h & Transformer & 80.87$\pm$1.23& 79.63$\pm$0.52\\
    \midrule
    BiomedCLIP & BiomedCLIP & Mamba & 77.04$\pm$0.74 & 76.19$\pm$2.69 \\
    MedImageInsight & MedImageInsight & Mamba & 76.13$\pm$0.53 & 77.04$\pm$1.53 \\
    MedImageInsight  & Prov-GigaPath & Mamba & 80.66$\pm$0.59 &  81.90$\pm$2.69\\
    MedImageInsight  & UNI2-h & Mamba & \textbf{81.27$\pm$0.56} & \textbf{85.71$\pm$5.23}\\
    \bottomrule
    \end{tabular}
    \end{adjustbox}
    \caption{Ablation Studies with Variable Healthcare Foundation Models and Adapter Network Architectures}
\end{table*}

\subsection{Ablation Studies with Current Foundation Models and Adapter Networks}
Beyond comparing our method to current multi-modal survival prediction approaches, we conducted an extensive ablation study (Table 2) to assess the impact of different adapter network architectures and foundation model embedding combinations. The ablation results suggests that both the choice of foundation model and the adapter architecture play a critical role in achieving optimal predictive performance. Notably, the Mamba adapter consistently outperforms other architectures by effectively capturing fine-grained cross-modal interactions, thereby validating the advantages of our integrated multi-modal framework.\\
\textbf{Foundation Models.} Our ablation study reveals that the choice of foundation models significantly impacts performance. When substituting a specialized encoder with a less targeted alternative—such as replacing MedImageInsight with BiomedCLIP for radiology—the performance consistently declines. This indicates that domain-specific pre-training, as seen in MedImageInsight, captures more nuanced and discriminative features that are essential for accurately characterizing glioma tumors. Similarly, using pathology-specific models like Prov-GigaPath and UNI2-h, lead to superior outcomes compared to more generic models. The combination of MedImageInsight for radiology and UNI for pathology delivers the highest c-index, suggesting that the tailored features learned from each modality provide critical insights for survival prediction. Essentially, models that are pre-trained on domain-specific data better capture the intrinsic patterns and clinical markers relevant to glioma, ultimately enhancing the predictive performance. \\
\textbf{Adapter Architectures.} The architecture of the adapter network used to fuse the modality-specific embeddings also plays a crucial role in overall performance. Across all tested combinations of foundation models, our proposed Mamba adapter consistently outperforms traditional fusion architectures, such as MLP and transformer-based adapters. For instance, when using MedImageInsight and UNI, the Mamba adapter improves the c-index from a range of $79.46\pm0.10$  to $81.27\pm0.56$. This notable improvement is attributed to the Mamba adapter's token-based attention mechanism, which is specifically designed to capture and preserve subtle cross-modal relationships. By treating each modality embedding as a token, the Mamba adapter facilitates dynamic interactions between modalities, leading to a more refined and comprehensive joint representation. This richer feature integration allows the network to exploit intricate dependencies that are often missed by simpler architectures, ultimately resulting in a more robust prediction.\\
\textbf{Survival Curve Analysis.} To further validate these findings, Figure 2 presents Kaplan–Meier survival curves for high-risk and low-risk groups as predicted by each ablated variant. Both configurations—MedImageInsight paired with either Prov-GigaPath or UNI2-h, and using the Mamba and transformer adapter—yield distinctly separated curves. Notably, the multi-modal embeddings with UNI2-h exhibit the best alignment with grade 0 and grade 2 subjects, while the configuration with Prov-GigaPath more effectively differentiates between grade 1 and other subjects. These qualitative observations are consistent with the quantitative improvements in the C-Index, underscoring the critical role of both the foundation model selection and the adapter architecture in accurately capturing survival outcomes.

\section{Conclusion}
In this paper, we introduce M4Survive, a novel multi-modal fusion approach that leverages foundation model embeddings to construct a unified semantic space, with the selective modeling of Mamba adapter. Our method achieves state-of-the-art performance in survival prediction, and extensive ablation studies confirm that both the selection of domain-specific foundation models and the design of the adapter network are critical for effective embedding fusion. For future work, we aim to benchmark our approach on larger datasets and further investigate the effectiveness of radiology embeddings derived from 3D foundation models.
\bibliographystyle{splncs04}
\bibliography{paper3863}

\end{document}